\documentclass[letterpaper, 10 pt, conference]{ieeeconf}
\IEEEoverridecommandlockouts
\usepackage{cite}
\usepackage{amsmath,amssymb,amsfonts}
\usepackage{bm}
\usepackage{algorithmic}
\usepackage{graphicx}
\usepackage{textcomp}
\usepackage{xcolor}
\usepackage{booktabs} 
\usepackage{scalerel,stackengine}
\usepackage[per-mode=symbol]{siunitx}
\stackMath
\newcommand\reallywidehat[1]{%
\savestack{\tmpbox}{\stretchto{%
  \scaleto{%
    \scalerel*[\widthof{\ensuremath{#1}}]{\kern-.6pt\bigwedge\kern-.6pt}%
    {\rule[-\textheight/2]{1ex}{\textheight}}
  }{\textheight}%
}{0.5ex}}%
\stackon[1pt]{#1}{\tmpbox}%
}

\usepackage{color}

\def\BibTeX{{\rm B\kern-.05em{\sc i\kern-.025em b}\kern-.08em
    T\kern-.1667em\lower.7ex\hbox{E}\kern-.125emX}}
\begin{document}

\title{The MIT Humanoid Robot: Design, Motion Planning, and Control For Acrobatic Behaviors}
\author{Matthew Chignoli\footnotemark$^{1}$, Donghyun Kim\footnotemark$^{2}$, Elijah Stanger-Jones\footnotemark$^{3}$, and Sangbae Kim\footnotemark$^{1}$  \thanks{$^{1}$Department of Mechanical Engineering and $^{3}$Department of Electrical Engineering and Computer Science, Massachusetts Institute of Technology, Cambridge, MA 02139, USA: {\tt\small chignoli@mit.edu}
$^{2}$College of Information and Computer Sciences, University of Massachusetts Amherst, Amherst, MA 01003, USA: {\tt\small donghyunkim@cs.umass.edu} }}


\maketitle
\thispagestyle{empty}
\pagestyle{empty}

\begin{abstract}
Demonstrating acrobatic behavior of a humanoid robot such as flips and spinning jumps requires systematic approaches across hardware design, motion planning, and control. In this paper, we present a new humanoid robot design, an actuator-aware kino-dynamic motion planner, and a landing controller as part of a practical system design for highly dynamic motion control of the humanoid robot. To achieve the impulsive motions, we develop two new proprioceptive actuators and experimentally evaluate their performance using our custom-designed dynamometer. The actuator's torque, velocity, and power limits are reflected in our kino-dynamic motion planner by approximating the configuration-dependent reaction force limits and in our dynamics simulator by including actuator dynamics along with the robot's full-body dynamics. For the landing control, we effectively integrate model-predictive control and whole-body impulse control by connecting them in a dynamically consistent way to accomplish both the long-time horizon optimal control and high-bandwidth full-body dynamics-based feedback. Actuators' torque output over the entire motion are validated based on the velocity-torque model including battery voltage droop and back-EMF voltage. With the carefully designed hardware and control framework, we successfully demonstrate dynamic behaviors such as back flips, front flips, and spinning jumps in our realistic dynamics simulation.
\end{abstract}


\newcommand{\Real}{\mathbb{R}}
\newcommand{\omegaBdHat}{\widehat{^B\bm{\omega}^d}}

\section{Introduction}

It has been three years since Boston Dynamics released the video of their humanoid robot, ATLAS \cite{atlas_backflip}, performing a backflip. However, no supplemental documentation was provided to answer questions regarding their methodology, the challenges they faced, or guidelines for replicating similar motions. We can speculate that the acrobatic motion is accomplished via a combination of their high powered robot and planning/control algorithms that are carefully designed based on the performance and limitations of their robot hardware. As an important step towards such demonstrations of highly dynamic motion, we present a new robot hardware design, planning, and control framework that offers practical solutions to the aforementioned questions. 

The 2015 DARPA Robotics Challenge (DRC) laid the groundwork for the next generation of humanoid robots~\cite{kuindersma2016optimization,kaneko2015humanoid,jung2018development}. The designs of these robots and the algorithms used to control them, however, have been primarily aimed at accomplishing the minimally dynamic tasks required for the DRC~\cite{dai2016planning,griffin2019footstep,feng2015optimization}. The most common design paradigms for these modern humanoid robots include hydraulically actuated designs~\cite{nelson2012petman,hyon2016design} and designs that utilize physically compliant actuators~\cite{radford2015valkyrie,tsagarakis2017walk}. These actuation paradigms must contend with issues like high mechanical impedance and limited force control bandwidth, respectively, that create significant challenges for achieving acrobatic motions. The proprioceptive actuator design of the MIT Cheetah robot’s has demonstrated a successful solution to this trade-off~\cite{wensing2017proprioceptive}, but to date has not been implemented on a humanoid robot.

\begin{figure}[t]
    \centering
    \includegraphics[width=\columnwidth]{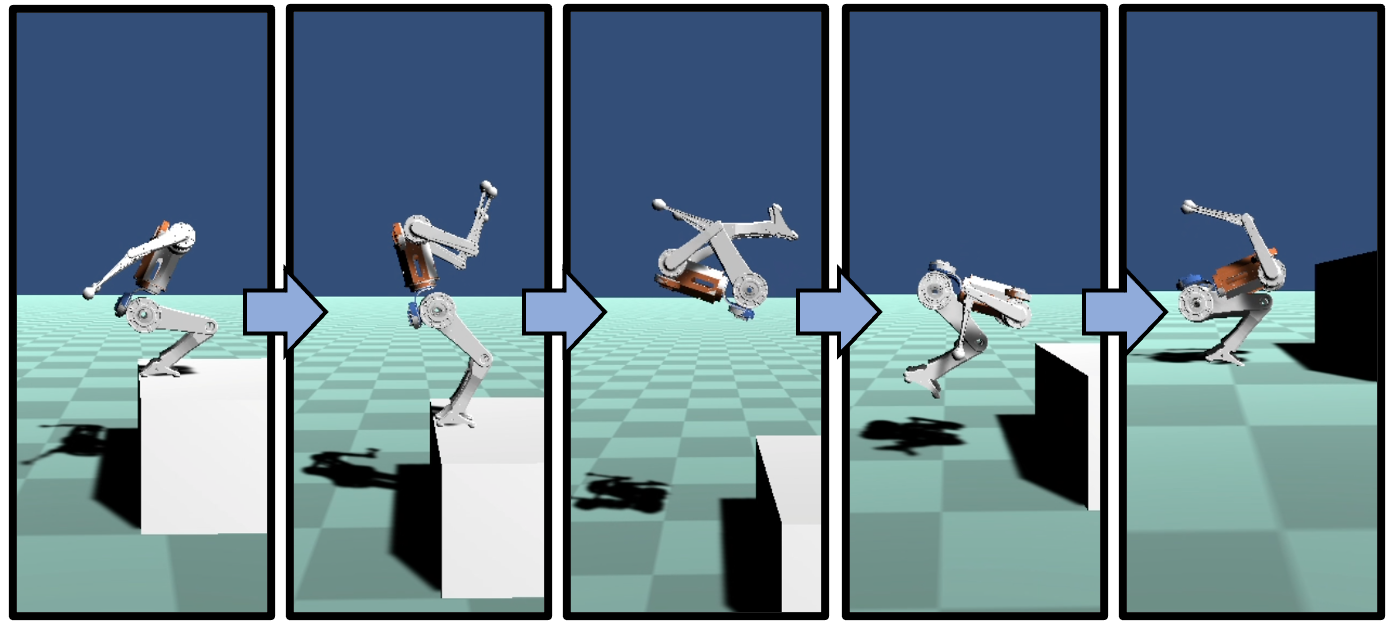}
    \caption{{\bf MIT Humanoid Backflip.} Proposed system designed is used to performing a back flip off of a humanoid robot off of a 0.4~\si{\meter} platform.}
    \label{fig:back-flip-1}
\end{figure}

Manually finding acrobatic motion trajectories is nearly impossible because of the complexity of the motions and the high degree of freedom (DoF) of a humanoid robot. Consequently, trajectory optimization-based methods have emerged as the most widely used solution for the problem. Trajectory optimization-based approaches to motion planning that use full-body dynamics~\cite{posa2014direct,mordatch2015ensemble} can exploit the full dynamic range of the robot, but are prone to issues like local minima and excessively long solve times. These issues can be circumvented by using a reduced-order model of the robot, like a spring-mass model~\cite{wensing2013high,hereid2014embedding}, but restrictive assumptions typically limit the generality of these approaches. 
Centroidal dynamics-based approaches to motion planning are popular due to their unique combination of computational tractability and dynamic expressiveness. By considering only the centroidal dynamics of the robot~\cite{orin2013centroidal}, these approaches are able to capture the core dynamics of the system without having to contend with the robot’s numerous degrees of freedom. Kino-dynamic planners~\cite{dai2014whole,herzog2016structured} simultaneously optimize the robot's centroidal dynamics and joint-level kinematics, which offer advantages in terms of generality.

For stable landing, the robot must dissipate kinetic energy over a long time period while rapidly responding to the error in the controller's expectation caused by unmodeled dynamics, modeling error, or external disturbance. This challenge has been addressed in our previous work \cite{kim2019highly} that proposed a hierarchical control framework integrating the model-predictive control (MPC) using a simple lumped-mass model for long-time horizon optimization and whole-body impulse control (WBIC) for instantaneous high bandwidth control. In the previous work, WBIC shared the same position command with MPC. Here, WBIC takes the optimal motion found in MPC as the position reference along with the reaction force commands to fully utilize the optimal solution. Furthermore, WBIC prioritizes a body orientation task over a centroidal momentum task to track desired centroidal angular momentum as long as the body orients command is not violated. With this carefully framed integration, we accomplish the stable landing of a humanoid robot on the ground after dynamic aerial motions. 
\begin{figure}
    \centering
    \includegraphics[width=0.9\columnwidth]{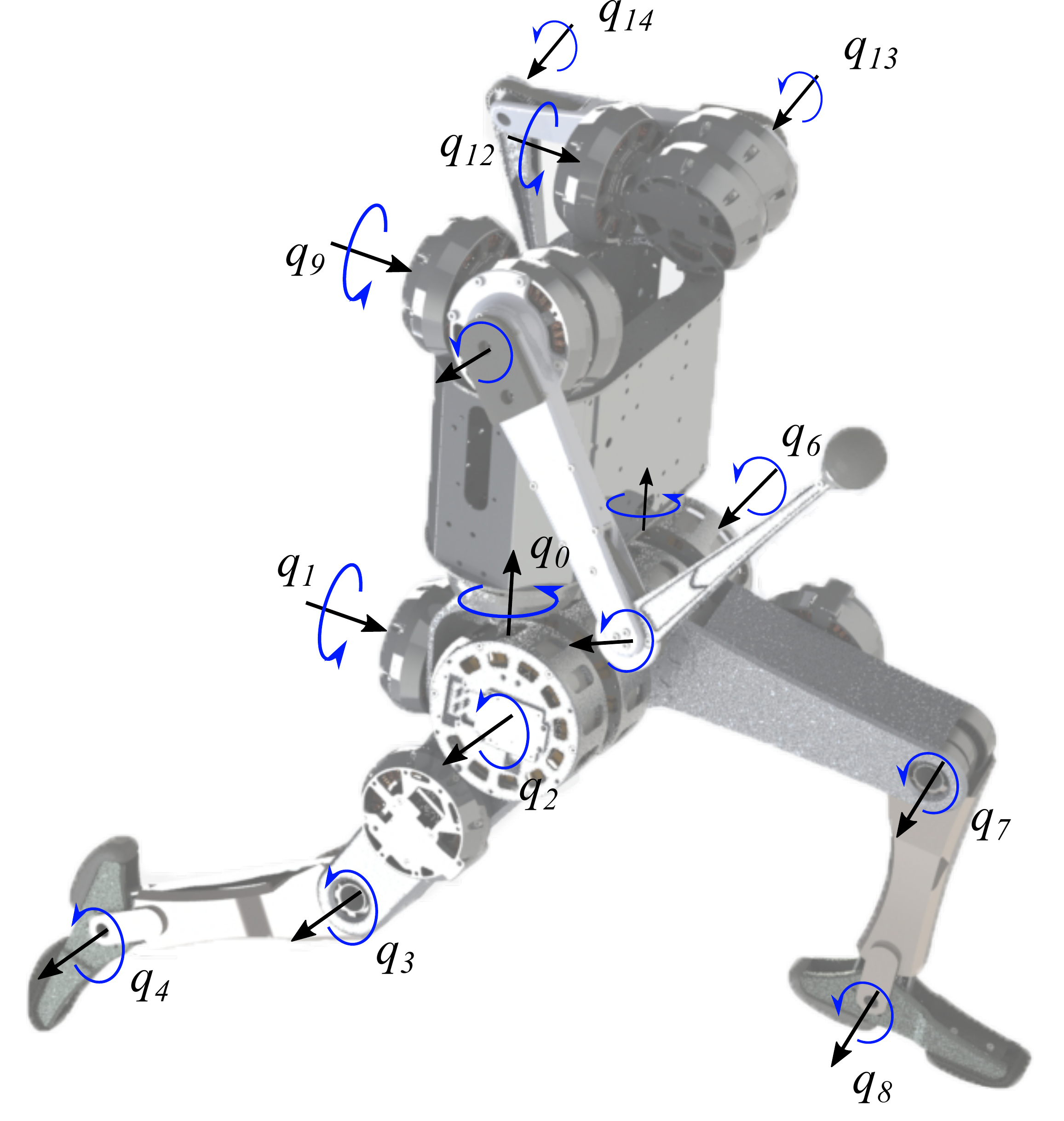}
    \caption{{\bf Configuration of the MIT Humanoid robot.} MIT humanoid has 5-DoF legs and 3-DoF arms. The ankle joint has only pitch directional actuation. Two contact sensors are located at toe and heel of each foot.}
    \label{fig:humanoid-configuration}
\end{figure}

The main contribution of this paper is a unified approach to design, motion planning, and control of a humanoid robot that enables highly dynamic motions like the back flip shown in Fig.~\ref{fig:back-flip-1}. The MIT Humanoid robot is the first attempt at applying the highly successful design principles of the MIT Cheetah robots~\cite{seok2013design,park2015variable,bledt2018cheetah,katz2019mini} to a humanoid robot. In order to leverage the full dynamic capabilities of the robot in the impulsive motions, a novel kino-dynamic planner is developed that efficiently deals with the actuator limits of the robot. 
For the stable landing the jumps, we utilize a hierarchical control framework by effectively integrating a model-predictive control and whole-body control. 

The viability of our proposed system design is demonstrated through simulation experiments of the robot performing acrobatic motions like flips, spins, and jumps. The performance of the MIT Humanoid's custom actuators is experimentally validated and included in the simulation experiments of the robot to ensure demonstrated motions will be feasible on the fully assembled robot.


\section{System Overview}


The design paradigm of previous MIT Cheetah robots has involved a unique combination of torque dense electric motors, high-bandwidth force control, and the ability to mitigate impacts through backdrivability. This design paradigm allows the robot to produce the impulses it needs to propel itself into air while also providing mechanical robustness that enables reliable control throughout the high speed impacts that occur when landing. These same design principles have been used in the design of the MIT Humanoid; a virtual model of the near-final design is shown in Fig.~\ref{fig:humanoid-configuration}. 

%
\subsection{Robot Design}
The MIT Humanoid will stand approximately 0.7~\si{\meter} tall and weigh approximately 21~\si{\kilo\gram}. Approximately 75.6~\si{\percent} of that mass is contained in the robot's torso, shoulder, and hip, 22.5~\si{\percent} in the robot's legs and the remaining 1.9~\si{\percent} in the robot's arms. This distribution is largely invariant to the robot's pose. The legs of the robot each have five of our custom actuators: three for the hip, one for the knee, and one for the pitch of the ankle. The knee, ankle and elbow joints contain a belt gearing system to deliver higher torques at these joints. The five degrees of freedom allow each leg to produce 3D ground reaction forces at each foot as well as moments about each foot's pitch and yaw axes. The actuator setup is detailed in Fig.~\ref{fig:humanoid-configuration}. The system will be powered from a high performance 60V, 3Ah battery pack. Four wireless contact sensors are attached on heel and toe of each foot to detect time of contact.  

\begin{table}[thb]
\centering
\caption{Actuator specifications for the MIT Humanoid.}
\label{tab:actuator}
\begin{tabular}{c|clll}
\multicolumn{1}{c|}{Joint} & \begin{tabular}[c]{@{}c@{}}Motor \\ Module\end{tabular} & \begin{tabular}[c]{@{}c@{}}Gear \\ Ratio \end{tabular} & \begin{tabular}[c]{@{}c@{}}Max \\ Torque\end{tabular} & \begin{tabular}[c]{@{}c@{}}Max Joint \\ Speed\end{tabular} \\ \hline
Hip - Yaw & U10 & 6.0 & 33.6\si{\newton\meter} & 55\si{\radian\per\second} \\ \hline
Hip - Ab/Ad &U10 & 6.0 & 33.6\si{\newton\meter} & 55\si{\radian\per\second}  \\ \hline
Hip - Flexion &U12 & 6.0 & 68.0\si{\newton\meter} & 45\si{\radian\per\second} \\ \hline
Knee &U12 & 12.0 & 136.0\si{\newton\meter} & 22.5\si{\radian\per\second} \\ \hline
Ankle &U10 & 9.33 & 52.2\si{\newton\meter} & 35\si{\radian\per\second} \\ \hline
Shoulder - Ab/Ad & U10 & 6.0 & 33.6\si{\newton\meter} & 55\si{\radian\per\second} \\ \hline
Shoulder - Flexion & U10 & 6.0 & 33.6\si{\newton\meter} & 55\si{\radian\per\second} \\ \hline
Elbow & U10 & 9.33 & 52.2\si{\newton\meter} & 35\si{\radian\per\second}
\end{tabular}
\end{table}

\subsection{Custom Actuator Design}
\begin{figure}[thb]
    \centering
    \includegraphics[width=0.9\columnwidth]{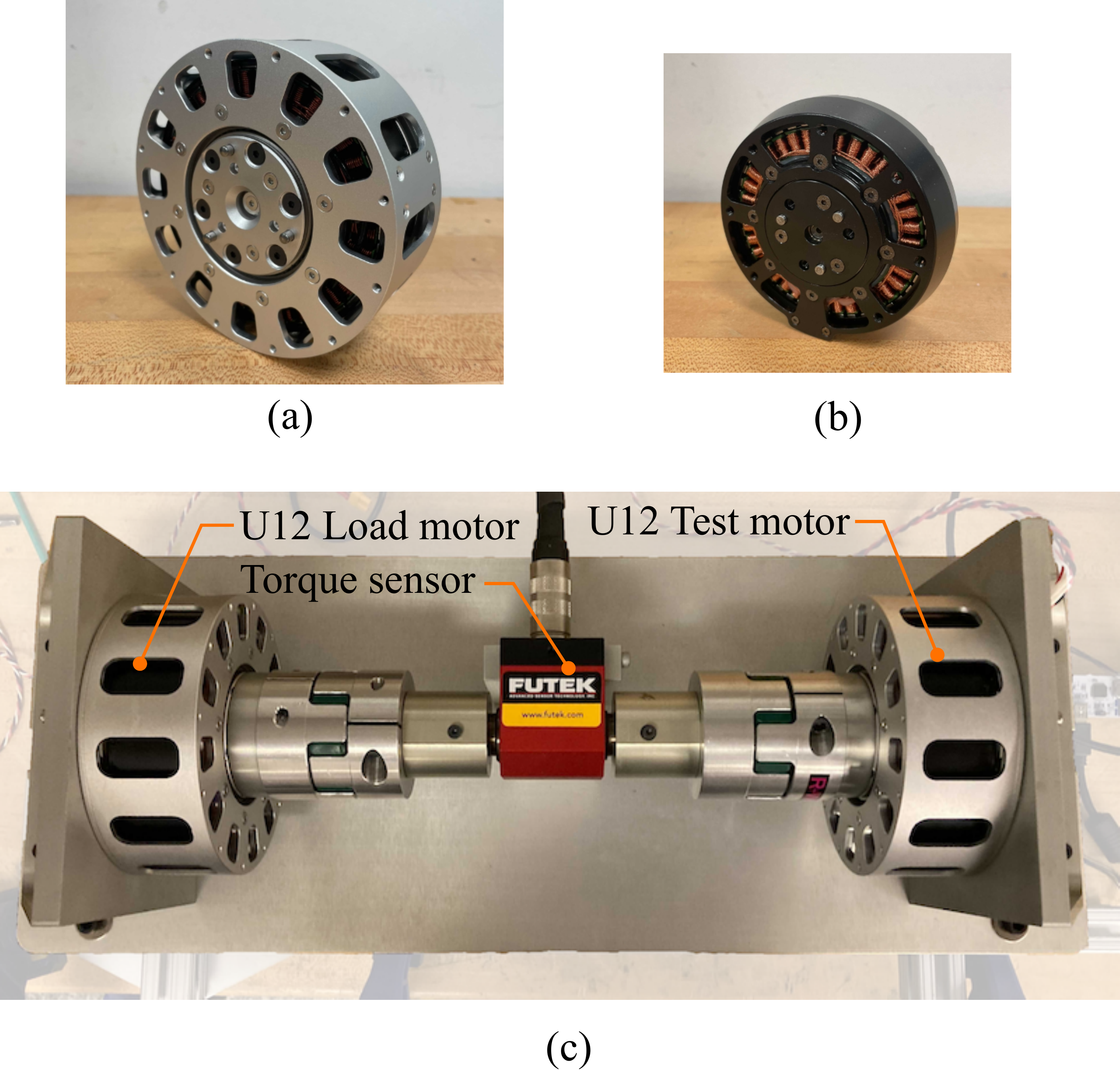}
    \caption{{\bf Actuator modules and dynamometer} (a) Full assembly of the U12 and (b) U10 motor modules. (c) our custom-design dynamometer.}
    \label{fig:actuator-design}
\end{figure}

The humanoid will use a combination of two custom, high torque density electric motor modules to actuate its joints. Both modules are based on the design of the actuators originally developed for the MIT Mini Cheetah~\cite{katz2019mini} but modified to meet the increased torque requirements of the MIT Humanoid. The details on the motor specifications can be found in Table \ref{tab:actuator}, and both modules are shown fully assembled in Fig.~\ref{fig:actuator-design}. Motor torque control is accomplished using a new version of the Mini-Cheetah motor drivers, modified to operate at the higher voltage and power levels necessary for this robot; there are no torque or force sensors. 

\begin{table}[thb]
\centering
\caption{Motor Module Design Specifications}
\label{tab:motorModule}
\begin{tabular}{c|ccc}
\multicolumn{1}{c|}{} & \begin{tabular}[c]{@{}c@{}}U10 \\ Module\end{tabular} & \begin{tabular}[c]{@{}c@{}}U12 \\ Module\end{tabular} \\ \hline
Mass (\si{\gram}) & 619 & 1174  \\ \hline
Gear ratio & 6:1 & 6:1  \\ \hline
Diameter (\si{\milli\metre}) & 101 & 120   \\ \hline
Thickness (\si{\milli\metre}) & 36 & 44   \\ \hline
Maximum Torque (\si{\newton\metre}) & 33.6 & 68.0  \\ \hline
Maximum Speed $@$60V  (\si{\radian\per\second}) & 55.0 & 45.0  \\ \hline
Output Inertia (\si{\kilogram\metre\squared}) & 0.0023 & 0.02
\end{tabular}
\end{table}

To validate the motors' capabilities and create an accurate model for our dynamics simulation, we constructed a custom dynamometer with a Futek TRS300 torque sensor.   

\begin{figure}[thb]
     \centering
     \includegraphics[width=\columnwidth]{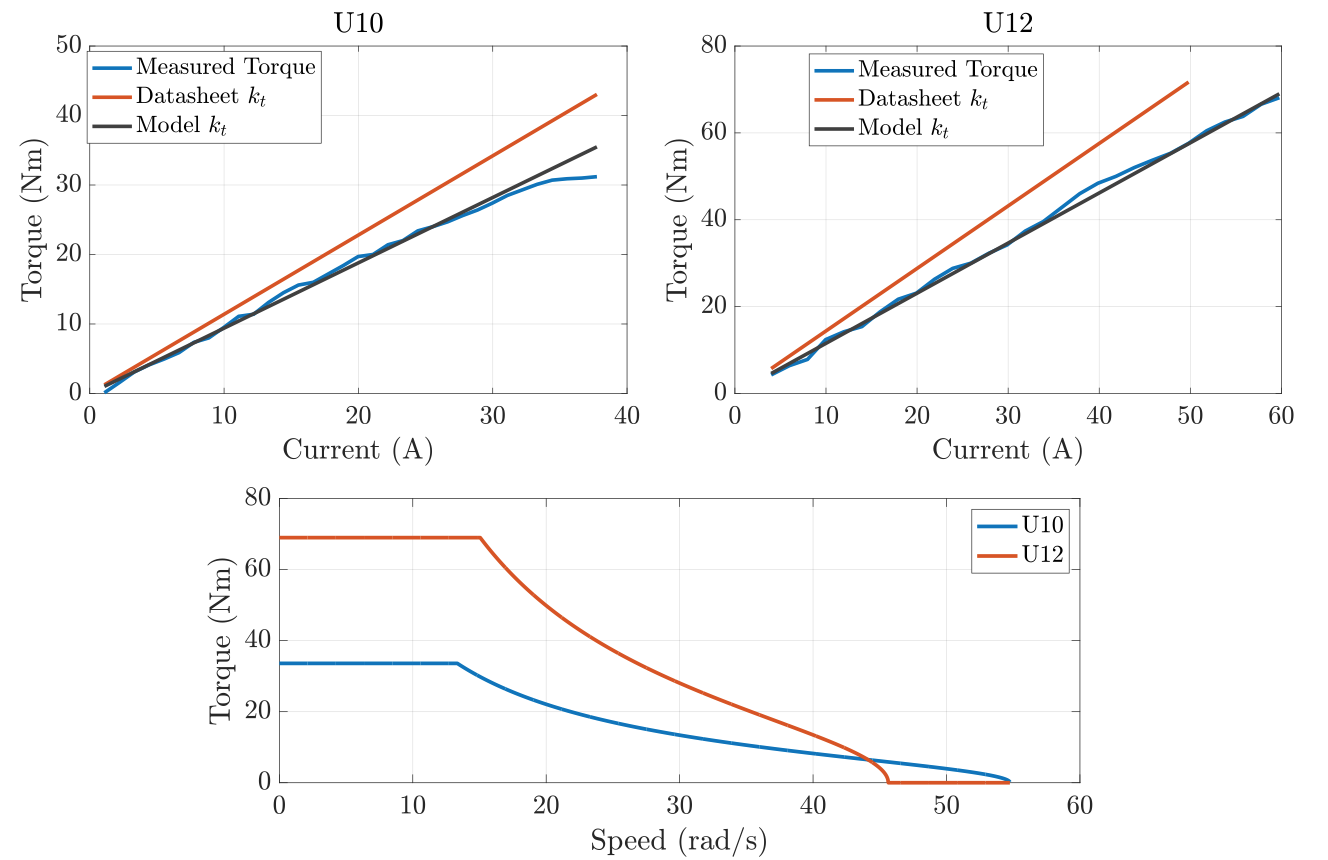}
     \caption{(Top) Current vs Torque for for the two motor modules. (Bottom) Empirically verified torque-speed curve for the two motor modules at 60V.}
     \label{fig:current-torque}
 \end{figure}

 The first step to verify the capabilities of the actuators is to measure peak torque and the torque constant using the dynamometer. Peak torque is the most critical value to ensure the model is capable of the dynamic motions in hardware. These results are detailed in Fig. \ref{fig:current-torque}. The U10 module reaches saturation at around 31Nm, however for the U12 saturation does not occur as the mechanical limits of the gearbox are reached first at 68Nm. 

\subsection{Actuator and Battery Dynamic Model}
 By combining the measured$k_t$ value with inductance and resistance measurements taken on an impedance analyzer we can create a highly accurate model of the motors torque speed curve. The torque produced by our motor is determined by
 \begin{equation}
     \tau = \frac{3}{2} p i_q \lambda,
 \end{equation}
 where $\tau$ is the produced torque, $p$ is the number of pole pairs, $i_q$ is the current on the $q$ axis, and $\lambda$ is the flux linkage.  
 However the motor cannot produce arbitrary torques as the maximum current that can be put into the motor is dependant on the stator voltage $V_s$, if this exceeds the current bus voltage due to back EMF at high speeds then $i_q$ will have to be limited, reducing torque capabilities. Note using a non-zero $i_d$ ($d$-axis current - known as field-weakening) can increase torque capabilities at a reduced efficiency, however this was not necessary to achieve the trajectories in this paper. Torque speed curves for any bus voltage can be generated using 
 \begin{eqnarray}  
     V_s &=& \sqrt{(R_s i_s)^2 + (\omega p)^2 \left((L_q i_q)^2 + (\lambda + L_d i_d)^2\right) }, \label{eq:ts} \\ 
     i_s &=& \sqrt{i_q^2 + i_d^2},
 \end{eqnarray}
where $R_s$ is stator resistance, $i_s$ is total stator current, $\omega$ is rotational velocity, $L_q$ and $L_d$ are $q$- and $d$-axis inductances.  The motor actuator model also includes estimates for damping and friction coefficients.
 
In previous work \cite{Katz:2019vk}, we have seen issues where the limitations of the on-board battery can cause differences between capabilities of simulation and reality; especially during highly dynamic movements. As a battery draws more power, its output voltage sags which limits the capabilities of the motors. Note voltage cannot drop further than 50~\si{\percent} of its fully charged state as this will greatly damage the battery. This phenomena is reflected in the relationship
\begin{eqnarray}
P &=& \sum_i^{n_m} \frac{\dot q_i \tau_i}{\eta},  \\
V_\text{fin} &=& \frac{1}{2} \left(V_\text{init} + \sqrt{V_\text{init}^2 - 4 P R_b}\right),
\end{eqnarray}
where $R_b$ is the measured battery impedance, $P$ is total power from the motors, $\eta$ is the estimated motor efficiency found from stator resistance, $n_m$ is the total number of motors, $V_\text{init}$ is the bus voltage at no load and $V_\text{fin}$ is the bus voltage at the given load. By combining this new voltage estimate with the torque speed curve calculations in \eqref{eq:ts} we can determine highly accurate actuator limits at every time step in simulation.   

\section{Planning and Control} \label{sec::kd}
\begin{figure}
    \centering
    \includegraphics[width=\columnwidth]{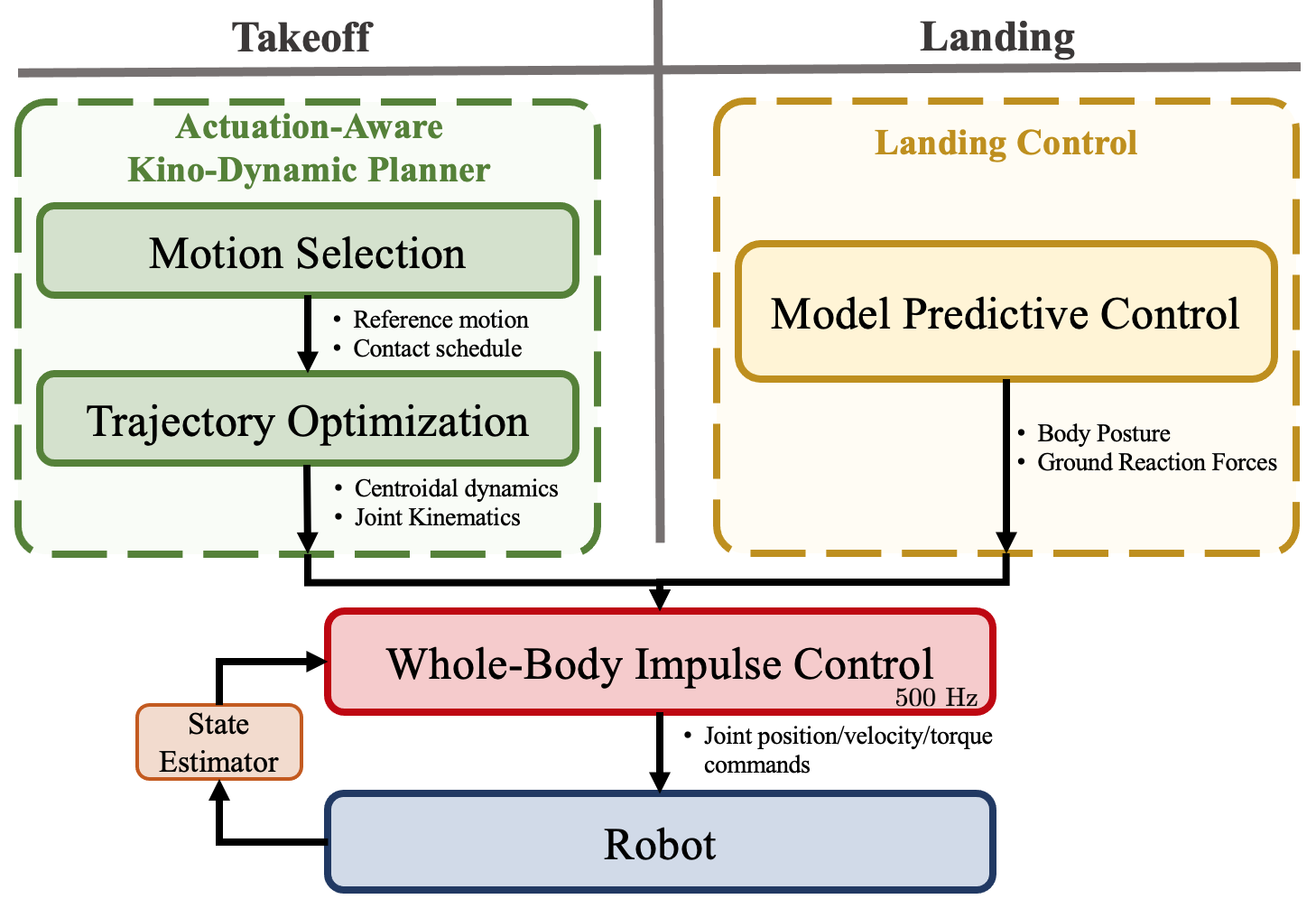}
    \caption{{\bf Planning and Control Framework.} Kino-dynamic planning, MPC-based landing control, and WBIC framework to achieve dynamic aerial motions of a humanoid robot.}
    \label{fig:framework}
\end{figure}

Dynamic aerial motions of the humanoid robot are achieved via the planning and control framework illustrated in Fig.~\ref{fig:framework}. The motions are broken down into three phases: takeoff, flight, and landing. For takeoff, planning is carried out via the novel Actuation-Aware Kino-Dynamic Planner (AAKD). The first stage of AAKD planning is a motion selector, which, based on an evaluation of the terrain and the task at hand, selects the necessary motion for the robot (e.g. jump onto the table, do a backflip, etc.). The second stage consists of a centroidal dynamics-based trajectory optimization that finds a dynamically feasible trajectory for the robot that accomplishes the selected motion. For flight, the joints of the robot are simply controlled using PD control to a configuration determined by the AAKD planner. For landing, a quadratic programming (QP)-based MPC is used to plan reaction force profiles to stabilize the robot as it comes back into contact with the ground. In both the takeoff and landing phases, planned motions are realized via a whole-body impulse controller optimized for stabilizing motions with large rotational components.

\subsection{Kino-Dynamic Planning}
The optimization variables of the kino-dynamic optimization for a robot with $n$ joint degrees of freedom include discrete trajectories for the generalized position vector of the robot $\mathbf{q}\in\Real^{n+6}$, the generalized velocity vector for the robot $\mathbf{\dot{q}}\in\Real^{n+6}$, the position $\mathbf{r}\in\Real^3$, velocity $\dot{\mathbf{r}}\in\Real^3$, and acceleration $\ddot{\mathbf{r}}\in\Real^3$ of the robot's center of mass (CoM), the robot's centroidal angular momentum (CAM) $\mathbf{h}\in\Real^3$, it's time derivative $\mathbf{\dot{h}}\in\Real^3$, the positions of the robot's contact points $\mathbf{c}_i\in\Real^3$, and the external ground reaction forces acting at the contact points $\mathbf{f}_i\in\Real^3$. All $n_c$ contacts are treated as points contacts. Each of these trajectories are discretized into $n_t$ timesteps where $n_t$ is the total number of timesteps and $\Delta t$ is the time between each timestep. These variables are condensed into a single optimization variable $\mathbf{X}\in\Real^{(2n+6n_c+27)\times n_t}$ where $\mathbf{X} = [\mathbf{X}_1,...,\mathbf{X}_{n_t}]$ and 
\begin{equation}
    \mathbf{X}_k = \begin{bmatrix} \mathbf{x}_k^{\top} & \mathbf{r}_k^{\top} & \dot{\mathbf{r}}_k^{\top} & \ddot{\mathbf{r}}_k^{\top} & \mathbf{h}_k^{\top} & \dot{\mathbf{h}}_k^T & \mathbf{c}_{k}^{\top} & \mathbf{f}_{k}^{\top} \end{bmatrix}^{\top}, \label{eqn::opt_var}
\end{equation}
with
\begin{equation}
    \mathbf{x} = \begin{bmatrix} \mathbf{q} \\ \dot{\mathbf{q}} \end{bmatrix}, \quad \mathbf{c} = \begin{bmatrix} \mathbf{c}_1 \\ \vdots \\ \mathbf{c}_{n_c} \end{bmatrix}, \quad \mathbf{f} = \begin{bmatrix} \mathbf{f}_1 \\ \vdots \\ \mathbf{f}_{n_c} \end{bmatrix}. \label{eqn::foot_pos_force}
\end{equation}

The motion selector provides the trajectory optimization with a reference motion $\mathbf{X}_\text{ref}\in\Real^{(2n+6n_c+27)\times n_t}$ and a set of allowable terminal states $\bm{\mathcal{X}}_\text{term}$. While the framework is amenable to more sophisticated approaches to motion selection, the motions demonstrated in this work all emerge from a simple motion selector that specifies a terminal state of the robot and then generates a reference motion by interpolating between the initial state and this final state. Examples of the terminal state of the robot could be that the robot ends up on top of an elevated surface, that it ends up on the other side of a wide gap, or that it rotates 360\si{\degree} about one of its axes.

The terminal condition $\mathbf{X}_{n_t}\in\bm{\mathcal{X}}_\text{term}$ ensures that the output of the kino-dynamic optimization achieves the desired motion. The reference motion is used to guide the optimization towards the desired motion via the objective function
\begin{equation}
    \min_{\mathbf{X}} \quad \sum_{k=1}^{n_t} \vert\vert\mathbf{X}_{k,\text{ref}} - \mathbf{X}_k\vert\vert^2_{\bm{Q}_{\mathbf{X}}},
\end{equation}
where $\bm{Q}_{\mathbf{X}}\in\Real^{(2n+6n_c+27)\times (2n+6n_c+27)}$ is a weight matrix.

The following constraints are standard for kino-dynamic optimization~\cite{dai2014whole} and are enforced at every timestep $k$ to ensure kinematic and dynamic feasibility of the generated motions. First, dynamical constraints encode the centroidal dynamics of the robot
\begin{equation}
    \begin{split}
        m\ddot{\mathbf{r}}_k &= \sum_{i=1}^{n_c}\mathbf{f}_{i,k} + m\mathbf{g} \\
        \dot{\mathbf{h}}_k &= \sum_{i=1}^{n_c}(\mathbf{c}_{i,k} - \mathbf{r}_k)\times \mathbf{f}_{i,k}, \\
        \mathbf{h}_k &= \bm{A}_\text{CAM}(\mathbf{q}_k)\dot{\mathbf{q}}_k
    \end{split}
\end{equation}
where $\bm{A}_\text{CAM}$ is the angular portion centroidal momentum matrix~\cite{wensing2016improved}. Continuity between timesteps is enforced via backward Euler integration. Kinematic feasibility is ensured via range of motion constraints on the joints as well as constraints that enforce agreement between kinematics and centroidal dynamics
\begin{equation}
    \begin{split}
        \mathbf{r}_k &= \bm{\gamma}_\text{CoM}(\mathbf{q}_k) \\ 
        \mathbf{c}_{i,k} &= \bm{\gamma}_{c_i}(\mathbf{q}_k) \quad \text{for } i = 1,2,...,n_c,
    \end{split}
\end{equation}
where $\bm{\gamma}_\text{CoM}$ and $\bm{\gamma}_{c_i}$ are functions that map the joint configuration of the robot to center of mass and end effector positions, respectively. The last of the standard kino-dynamic constraints are the contact constraints. Because the contact schedule is specified prior to the optimization, constraints can be separately imposed for limbs in contact versus out of contact. Limbs in contact are subject to the constraints
\begin{equation}
    \begin{split}
        \mathbf{c}_{i,k} &= \mathbf{c}_{i,k-1}, \\
        c_{i,z} &= \bm{\gamma}_\text{gnd}(c_{i,x},c_{i,y}), \\
        \mathbf{f}_{i,k} &\in \bm{\mathcal{F}},
    \end{split}
\end{equation}
where $\bm{\gamma}_\text{gnd}$ returns the height of the ground at a given point and $\bm{\mathcal{F}}$ represents the friction cone or, in the case of this work, a linear approximation of it. Limbs out of contact are subject to
\begin{equation}
        c_{i,z} \ge \bm{\gamma}_\text{gnd}(c_{i,x},c_{i,y}).
\end{equation}

A novel aspect of the proposed kino-dynamic planner is the inclusion of constraints that ensure the actuator limits of the robot are respected. Joint torques are not explicitly included in centroidal dynamics, and thus they do not show up in the kino-dynamic optimization. Mapping ground reaction forces to joint torques introduces significant nonlinearity to the formulation. For slow-moving motions like walking, actuator limits are typically dealt with via cost functions that penalize large reaction forces or constraints on maximum reaction force. In the case of highly dynamic motions that push the hardware limits of the robot, these simple techniques are not sufficient. 

At a given time step, the exact joints torques required for the commanded motion are given by
\begin{equation}
   \bm{\tau}_j = \bm{S}_j\left(\bm{H}\ddot{\mathbf{q}} + \bm{C} \dot{\mathbf{q}} +\bm{\tau}_G -\sum_{i = 1}^{n_c} \bm{J}_{c,i}^{\top}\mathbf{f}_{c,i}\right), \label{eq:joint-torque}
\end{equation}
where $\bm{S}_j\in\Real^{n\times n+6}$ is a selection matrix for the joint dynamics of the system, $\bm{\tau}_j\in\Real^{n}$ is the vector of joint torques, $\bm{H}\in\Real^{n+6\times n+6}$ is the mass matrix, $\bm{C}\dot{\mathbf{q}}\in\Real^{n+6}$ is the Coriolis vector, and $\bm{\tau}_G\in\Real^{n+6}$ is the gravity vector~\cite{featherstone2014rigid}.  Empirically, we find that the reaction force terms dominate this relationship, and thus the required joints torques can be well approximated by
\begin{equation}
   \bm{\tau}_j \approx \bm{S}_j\left(-\sum_{i = 1}^{n_c} \bm{J}_{c,i}^T\mathbf{f}_{c,i}\right). \label{eq:joint-torque-approx}
\end{equation}
Even with this approximation, considerable nonlinearity remains due to the dependence of the contact Jacobian on the robot's configuration. Observations of generated acrobatic motions show that the contact Jacobians of the leg are minimally sensitive to the positions of the hip yaw joints, the hip ab/ad joint, and the ankle joint. To further approximate~\eqref{eq:joint-torque-approx}, the contact Jacobians of the leg are linearized with respect to a reference position of these three joints (no linearization occurs with respect to the hip flexion or knee joints). This approximate relationship between ground reaction forces and joint torques is used in a constraint that encodes a linear approximation of the torque speed relationship in Fig~\ref{fig:current-torque}.

\subsection{Landing Model-Predictive Control}
Once the robot detects full contact with the ground, the landing controller is activated and starts stabilizing the robot's balance by properly absorbing the impact and dissipating the kinetic energy. This requires high-bandwidth feedback control that can account for optimal motion over the long sequence of the landing motion. To accommodate both real-time computation requirement and long time period optimization, we utilize the hierarchical control framework proposed in \cite{kim2019highly} consisting of MPC and WBIC.  

In this MPC formulation, a simplified lumped mass model is used to speed up the optimization process. To simplify the formulation even further, we use linear time-invariant dynamics based on the assumption that the body orientation does not change much and use Euler angle for the body orientation state representation. The Euler angle has a singularity issue, but for simplicity, we follow the convention used in \cite{kim2019highly}. Since the body orientation is assumed to be nearly constant during the MPC horizon, the angular velocity and inertia tensor are defined as follows.
\begin{align}
     \dot{\bm{\Theta}}_t&\approx \bm{R}_0\bm{\omega}_{t}, \\
    _\mathcal{G}\bm{I} &\approx \bm{R}_0 \ _\mathcal{B}\bm{I} \bm{R}_0^{\top},
\end{align}
where $\dot{\bm{\Theta}} = \begin{bmatrix} \dot{\phi} & \dot{\theta} & \dot{\psi} \end{bmatrix}^{\top}$ is roll ($\phi$), pitch ($\theta$), and yaw ($\psi$) velocity of the body, $\bm{R}_0\in SO(3)$ is a current body orientation, which translates the angular velocity in the global frame, $\bm{\omega}\in\Real^3$, to the local (body) coordinate, and $_\mathcal{G}\bm{I}\in\Real^{3 \times 3}$ and $_\mathcal{B}\bm{I}\in\Real^{3 \times 3}$ are the inertia tensor seen from the global and local (body) frame, respectively.

The final approximation used in the MPC is that one directional angular velocity is significant and off-diagonal terms of the inertia tensor are small. In most flip or turning jump cases, angular velocity about one principal axis is significantly larger than the others. For example, the pitch directional angular velocity is significant in the case of back/front flips and the yaw directional angular velocity is dominant in 180\si{\degree} turning jump. Based on this approximation, we ignore precession and nutation effects, $ \bm{I}\dot{\bm{\omega}} + \bm{\omega}\times\left( \bm{I} \bm{\omega} \right) \approx \bm{I}\dot{\bm{\omega}}$.

With the above simplifications, the discrete dynamics of the system can be expressed as
\begin{equation}
    \mathbf{x}(k+1) = \bm{A}_k\mathbf{x}(k) + \bm{B}_k\hat{\mathbf{f}} (k)+ \hat{\mathbf{g}},
\end{equation}
where,
\begin{equation}
\begin{split}
\mathbf{x} &= \begin{bmatrix}
\bm{\Theta}^{\top} & \mathbf{p}^{\top} & \bm{\omega}^{\top} & \dot{\mathbf{p}}^{\top} 
\end{bmatrix}^{\top}, \\[1.5mm]
    \hat{\mathbf{f}} &= \begin{bmatrix}
    \mathbf{f}_1 & \cdots & \mathbf{f}_n
    \end{bmatrix}^{\top},\\[1.5mm]
    \hat{\mathbf{g}} &= \begin{bmatrix}
    \bm{0}_{1\times3} & \bm{0}_{1\times3} & \bm{0}_{1\times3}&\mathbf{g}^{\top}
    \end{bmatrix}^{\top},
    \end{split}
\end{equation}
\begin{equation}
\begin{split}
    \bm{A} &= \begin{bmatrix}
    \bm{1}_{3\times3} & \bm{0}_{3\times3} & \bm{R}_0 \Delta t & \bm{0}_{3\times3} \\[1mm]
    \bm{0}_{3\times3} & \bm{1}_{3\times3} & \bm{0}_{3\times3} &
    \bm{1}_{3\times3} \Delta t \\[1mm]
    \bm{0}_{3\times3} & \bm{0}_{3\times3} & \bm{1}_{3\times3} &
    \bm{0}_{3\times3} \\[1mm]
    \bm{0}_{3\times3} & \bm{0}_{3\times3} & \bm{0}_{3\times3} &
    \bm{1}_{3\times3}
    \end{bmatrix}, \\[1.5mm]
    \bm{B} &= \begin{bmatrix}
    \bm{0}_{3\times3} & \cdots &  \bm{0}_{3\times3} \\[1mm]
    \bm{0}_{3\times3} & \cdots &  \bm{0}_{3\times3} \\[1mm]
_\mathcal{G}\bm{I}^{-1}[\mathbf{r}_1]_{\times} \Delta t& \cdots & _\mathcal{G}\bm{I}^{-1}[\mathbf{r}_n]_{\times} \Delta t\\[1mm]
    \bm{1}_{3\times3} \Delta t/m& \cdots & \bm{1}_{3\times3}
\Delta t/m    \end{bmatrix}.
\end{split}
\end{equation}
In the QP, we minimize the reaction forces and the error between the reference and state during the MPC horizon.
\begin{equation}\label{eq:mpc_qp}
    \min_{\mathbf{x}, \mathbf{f}} \sum_{k=0}^{m}||\mathbf{x}(k+1) - \mathbf{x}^{\rm ref}(k+1)||_{\bm{Q}} + ||\mathbf{f}(k)||_{\bm{R}}
\end{equation}
subject to dynamics, initial condition constraints, and the ground reaction force constraints, 
\begin{equation}
\label{eq:contact_constraint}
\mid f_x \mid \leq \mu f_z, \quad \mid f_y \mid \leq  \mu f_z,
\quad f_z > 0,
\end{equation}
where $\mu$ is the friction coefficient. The desired CoM position is defined by the center of the four contact locations of the feet (the right toe/heel and the left toe/heel) and the desired height is 0.05~\si{\meter} higher than the CoM height at the landing. The desired body orientation is also determined based on the landing posture but the pitch angle is adjusted to be smaller than 0.8~\si{\radian}. 
The time step used in MPC is 0.1 \si{\second} and the length of horizon is 15. Therefore, MPC optimizes 1.5~\si{\second} future state in one iteration. The update frequency of MPC is 10 Hz, which is slower than our real-time control frequency of 500~\si{\hertz}, while  WBIC runs at the real-time frequency to obtain high-bandwidth posture control.  

\subsection{Task Setup of Whole-Body Impulse Control}
The full formulation of WBIC is explained in \cite{kim2019highly}, and its overall framework is similar to other WBCs but has a feature to include the reference reaction forces coming from the planner or the MPC, which we explained in the previous sections. The performance of WBIC has been validated by showing the high-speed running of a quadruped robot, but two new important issues arise in the landing control. Note that these issues are specific to case where WBIC is used for landing control because, unlike the centroidal dynamics model using by the AAKD planner, the lumped-mass model used in MPC does not agree with the WBIC model in a straightforward way.

One issue is defining the orientation task. In the MPC, the orientation is simply defined by the orientation of the lumped mass. However, in the case of the humanoid robot, the body orientation has a different meaning than the lumped mass orientation because the body does not include all other link inertias. The most reasonable choice is the orientation of centroidal inertia \cite{orin2013centroidal}, however, controlling the angle of eigenvectors of the inertia does not exactly coincide with the output of MPC and difficult to add as a WBIC task. Fortunately, the mass of MIT Humanoid is concentrated around its body, and the actuators that dominante to robot's mass do not change location or orientation much while the robot moves. Therefore, we can approximate that the body orientation represents the orientation of the lumped mass and use that orientation as the reference of the body orientation task in WBIC. 

Another issue is selecting the task priorities. More specifically, the place of centroidal momentum (CM) task and how to define the task are the main question. The linear portion of CM is straightforward since it is simply CoM motion control, but the angular part is not easy as we partially described above. Minimizing CAM is desirable in most cases but not always since sometimes a robot must rotate its body to ensure the overall balance in the long-term perspective. Therefore, we take the orientation reference computed in MPC as a command of the body orientation as we mentioned above. The body orientation task is selected as the first task in the task prioritization. We locate the CM task under this task. The feedback law for the CM task does not include orientation error correction, but only centroidal angular velocity tracking. By prioritizing the tasks like this, we track the full-body angular momentum as long as the robot's orientation follows the reference input. 

To formulate null space projection of CM task, we need to define the Jacobian of the CM. \cite{4209816} suggested the Jacobian for CM inertia shaping, but in this work, we care more about CM motion than the shape of its inertia. Therefore, we use the Jacobian defined in \cite{7803411}, $J_{\rm CM}  = I_{\rm CM}^{-1}A_{CM}$. The other tasks below the CM task such as the joint posture task recognize the CAM minimization intention since the tasks are projected onto the null-space of the CM task, and automatically find an arm motion to counteract the body orientation control, which you can see in our resultant motions. 

\section{Results} \label{sec:results}
The simulation environment we use for testing the proposed system design is the custom dynamics simulator designed by the MIT Biomimetic Robotics Lab to address actuator dynamics as well as multi-DoF rigid-body dynamics of the robot. The performance of the simulator is extensively tested in our previous work using Mini-Cheetah robot, and the same framework for simulation actuator and multi-body dynamics is used for humanoid simulation experiments. Specifically, the model of the simulated actuators includes the effect of rotor inertias on the robot's dynamics, the torque-speed relationships of the motors, and the effect of battery voltage droop.

\begin{figure}[t]
    \centering
    \includegraphics[width=\columnwidth]{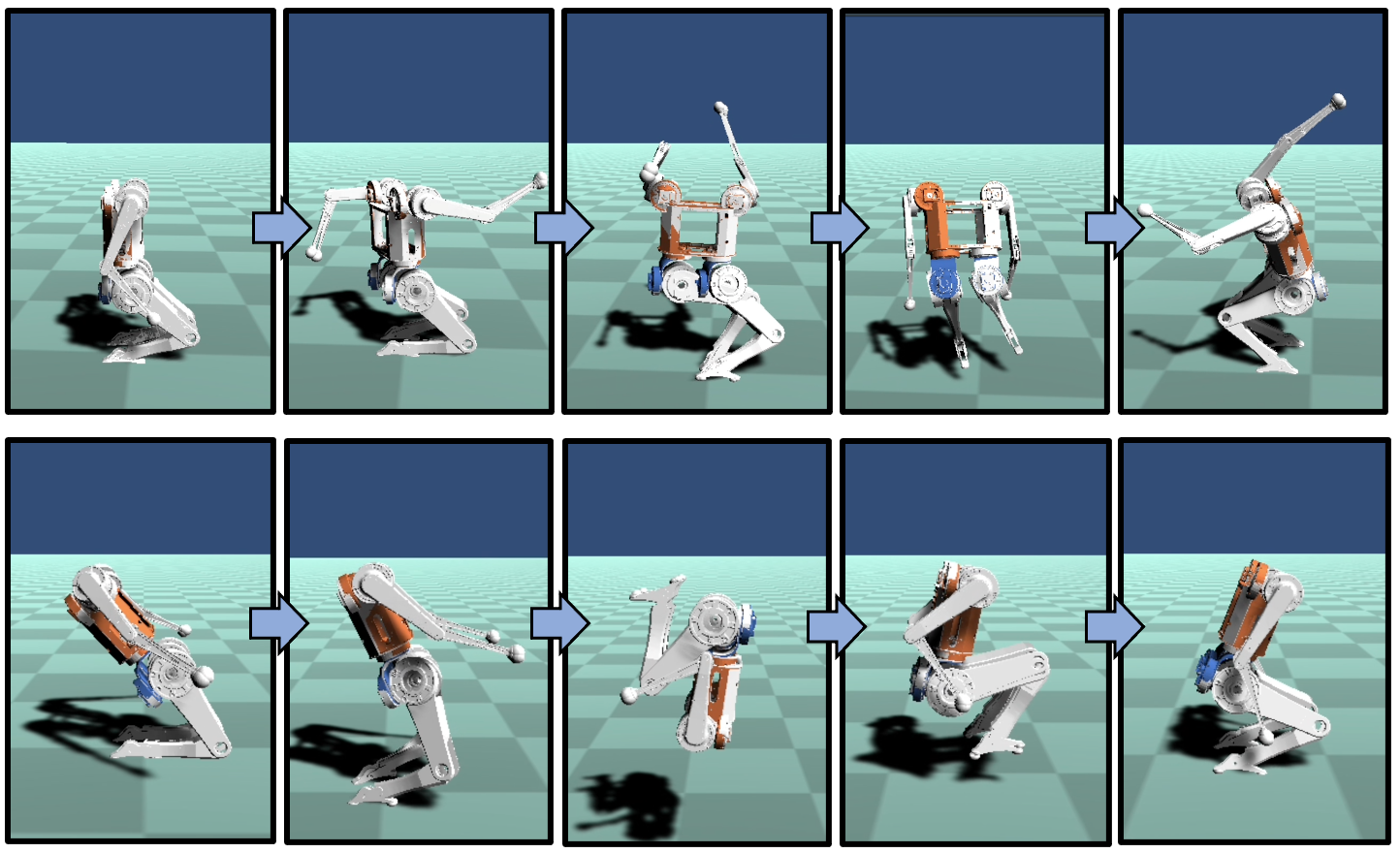}
    \caption{{\bf Acrobatic Humanoid.} MIT Humanoid performing a 180\si{\degree} spinning jump (top) and a standing front flip (bottom)}
    \label{fig:flip-spin}
    \vspace{-1mm}
\end{figure}

The dynamic capabilities of the humanoid robot are showcased through demonstrations of acrobatic abilities such as aerial flips and spins. While these acrobatic motions are not explicitly useful for legged robots, they serve as an important benchmark for what behaviors will be possible on the robot as future controllers are developed (e.g. parkour). All motion plans are generated using the AAKD planner discussed in Section~\ref{sec::kd}. The combination of terminal constraints and consideration of the robot's actuator limits allows trajectories to be rapidly generated and ready for immediate deployment on the robot. A comparison of the optimized joint torque profiles for standard kino-dynamic (KD) planning of a 180\si{\degree} spinning jump versus AAKD planning for the same jump shown in Fig.~\ref{fig:flip-spin}. This jumps exhibits relatively large displacements of the hip yaw and ab/ad joints, making it a worst-case scenario for the contact Jacobian approximation~\eqref{eq:joint-torque-approx}. Despite this, all actuator limits are respected in the AAKD case, although they are not respected in the KD case. The results in Table~\ref{tab:t-solve} compare the solve times for various acrobatic motions and show that AAKD only increasing solve times with respect to KD by an average of 28.2$\%$\si{\second}

\begin{figure}[t]
    \centering
    \includegraphics[width=\columnwidth]{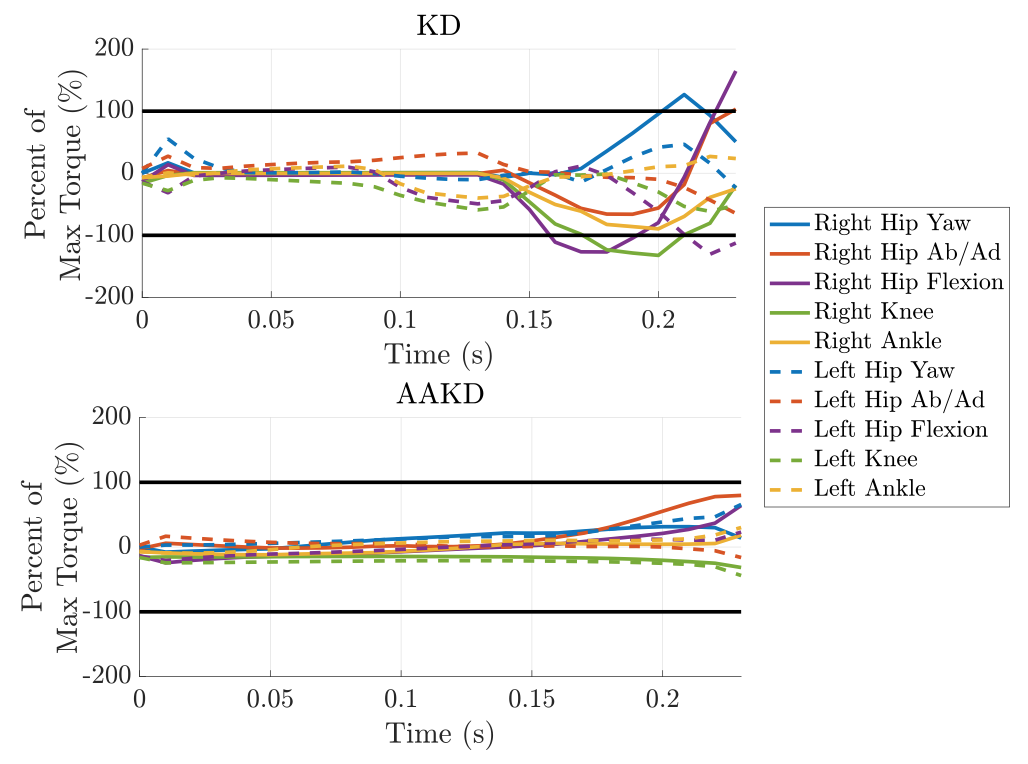}
    \caption{{\bf Actuation-Aware Planning.} Optimized joint torque profiles for a standing front flip of the MIT Humanoid planned with (top) a standard kino-dynamic planner (top) and (bottom) the proposed AAKD planner. Note that these profiles are only for the takeoff portion of the jump, landing and flight are not included.}
    \label{fig:TO-torque}
\end{figure}

\begin{table}[thb]
\centering
\caption{Solve Times for Kino-Dynamic Planners}
\label{tab:t-solve}
\begin{tabular}{|c|c|c|c|c|c|}
\hline
 & \begin{tabular}[c]{@{}c@{}}180\si{\degree}\\ Spin\end{tabular} & \begin{tabular}[c]{@{}c@{}}Forward\\ Jump\end{tabular} & \begin{tabular}[c]{@{}c@{}}Lateral\\ Jump\end{tabular} & \begin{tabular}[c]{@{}c@{}}Front\\ Flip\end{tabular} & \begin{tabular}[c]{@{}c@{}}Back\\ Flip\end{tabular} \\ \hline
KD & 4.09\si{\second} & 2.43\si{\second} & 2.98\si{\second} & 3.35\si{\second} & 6.51\si{\second} \\ \hline
AAKD & 4.67\si{\second} & 3.10\si{\second} & 3.24\si{\second} & 5.29\si{\second} & 8.77\si{\second} \\ \hline
\end{tabular}
\vspace{-2mm}
\end{table}

The AAKD planner produces motion that can coordinate high speed vertical propulsion of the robot's CoM with large angular momentum generation about the robot's principal axes to create the acrobatic motions demonstrated in this work, a few of which are shown in Fig.~\ref{fig:flip-spin}. Notice in these sequential screenshots that arm motions are used to stabilize the posture of the body. These arm motions are not authored by the AAKD planner, but rather, they emerge from the centroidal momentum task of the WBIC.

\begin{figure}[t]
    \centering
    \includegraphics[width=\columnwidth]{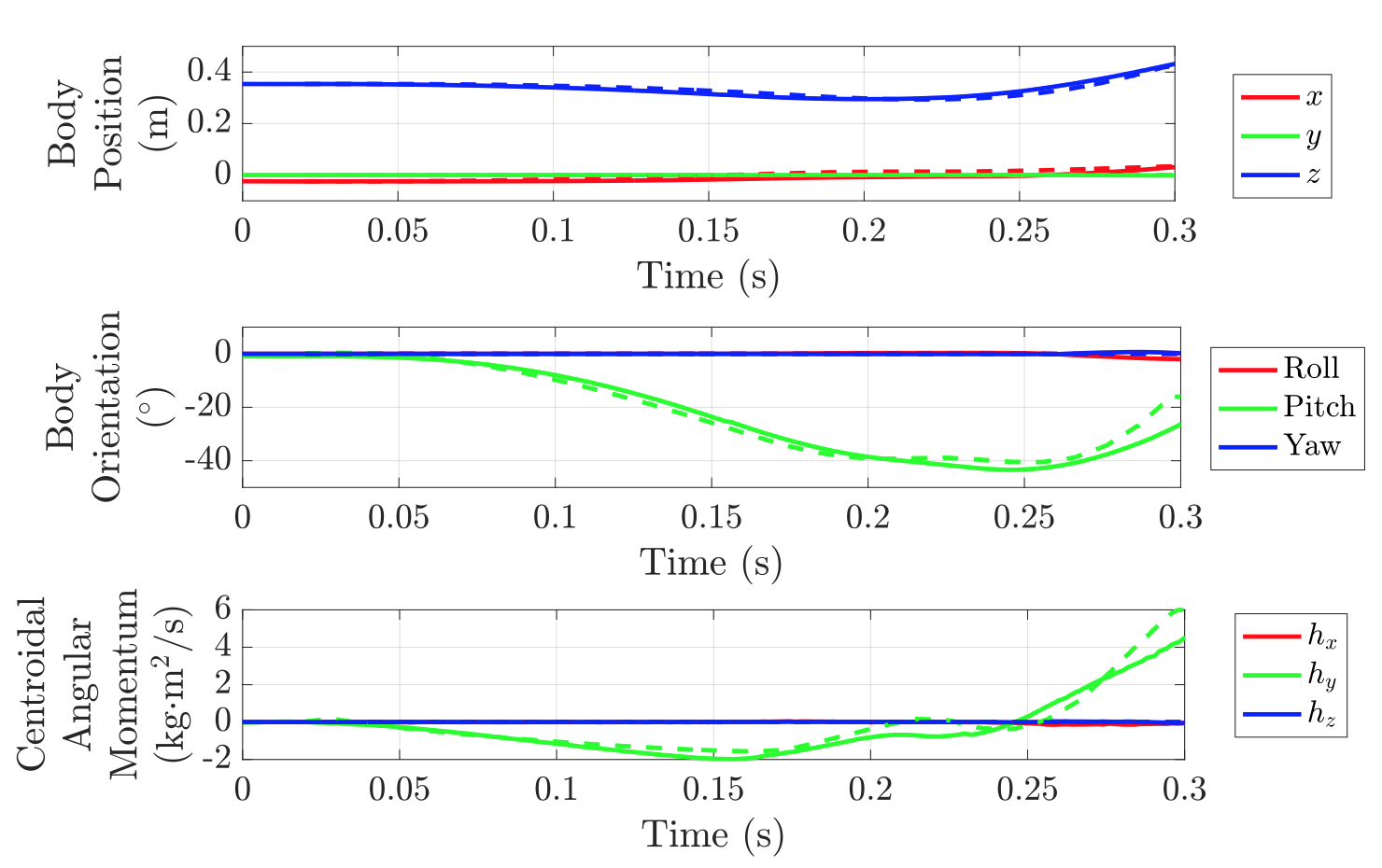}
    \caption{{\bf WBIC for Takeoff.} WBC tracking performance for the takeoff portion of the standing front flip of the MIT Humanoid.}
    \label{fig:front-flip-wbc}
    \vspace{-1mm}
\end{figure}

The high bandwidth full-body control offered by the WBIC makes it a useful for both the takeoff and landing phases of the demonstrated dynamic motions. The WBIC is able to account for the errors in the kino-dynamic and lumped-mass models of the robot as well as recover from small perturbations to nominal trajectories. An example of the tracking performance of the WBIC during the takeoff phase of the front flip is shown in Fig.~\ref{fig:front-flip-wbc}. The front flip is focused on in the discussion of the results because it involves the most significant rotation and the most severe impact with the ground, making it the worst-case scenario for both the takeoff and landing portions of the control framework. All other dynamic motions demonstrated in the attached video likewise exhibit robust tracking and obey all actuator limits.

\begin{figure}[t]
    \centering
    \includegraphics[width=\columnwidth]{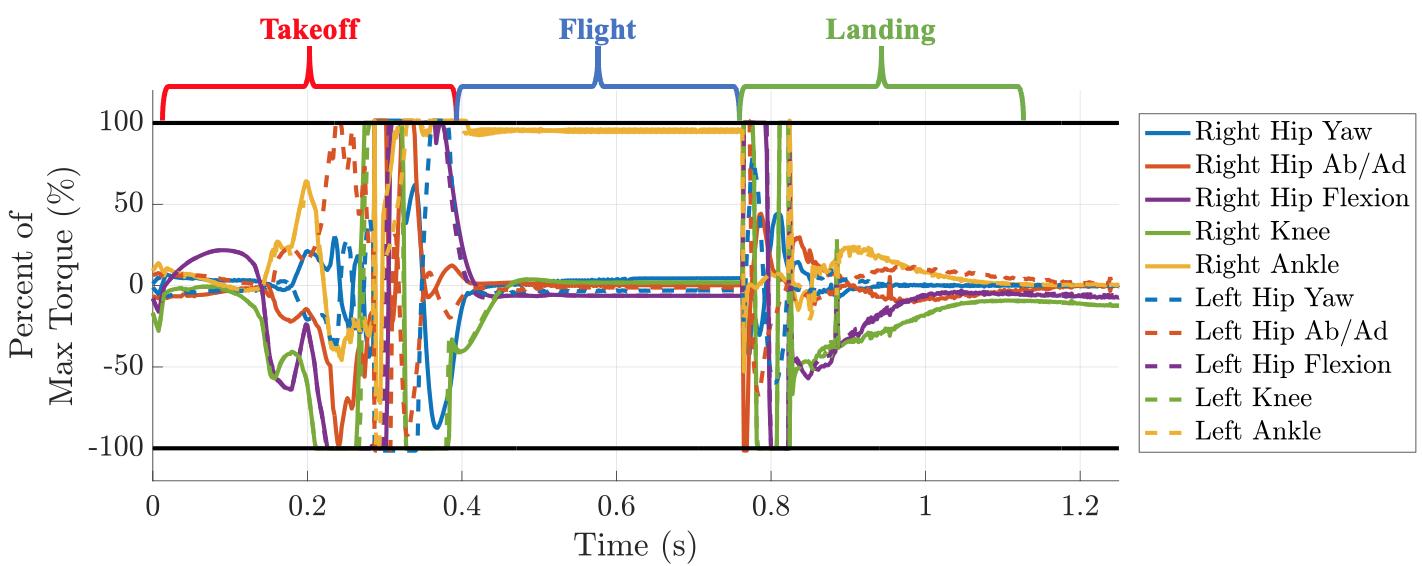}
    \caption{{\bf Simulated Actuator Model.} Simulated torque profiles of the MIT Humanoid performing a front flip.}
    \label{fig:sim-torque}
    \vspace{-2mm}
\end{figure}

While the acrobatic motions presented in this work are strictly simulation results, we again emphasize that extraordinary care is taken to ensure that the simulated dynamics reflect the true dynamics of the robot hardware. First, we include the inertia of each of the rotor inertia and account for the non-trivial effects that these spinning rotors have on the dynamics of the robot. We enforce that the simulated actuators obey the empirically verified torque-speed relationships illustrated in Fig.~\ref{fig:current-torque}. Lastly, we verify that the effects of battery voltage droop are reflected in the torque limits of the robot. The joint torque data in Fig.~\ref{fig:sim-torque} shows that the simulated torques exerted by the robot throughout all phases of motion virtually never exceed the limits imposed by our empirically verified actuator model. In these results, the experimental limit does not account for the effect of field weakening, so in reality the limit will be higher in many areas.

Since the actuators have been modeled great detail, only the following assumptions about the simulated system remain. First, we assume our model is accurate in terms of inertial properties. Second, we assume our rigid contact model of the ground is suitably accurate. Lastly, we assume the robot can adequately estimate its state. If these assumptions hold, which they likely will based on previous experiments with the MIT Cheetah robots, then we can be nearly certain that all demonstrations shown in this paper are ready for hardware implementation.

\section{Conclusions}
In this paper, we present a system design for the acrobatic motion demonstration of a humanoid robot by proposing a new humanoid robot, motion planner, and landing controller. Two new actuators are developed to satisfy the torque and power requirement for impulsive behaviors. On the software side, we formulate a new torque limit constraint that can be integrated into the simplified model based kino-dynamic planner to find a feasible trajectory in a real robot. The optimal trajectories are simulated in our custom dynamics simulator including actuator models and the resultant actuator outputs are verified again with a complete torque-velocity actuator model including battery voltage droop and back-EMF effect. For landing, we integrate MPC and WBIC with special care for dynamical consistency in transferring the optimal output of MPC to WBIC orientation control.

\section*{Acknowledgments}
This work was supported by the Centers for ME Research and Education at MIT and SUSTech, Naver Labs, and the National Science Foundation Graduate Research Fellowship Program under Grant No. 4000092301.

\bibliographystyle{./bibliography/IEEEtran}
\bibliography{chignoliHumanoid2020.bib}

\end{document}